\documentclass[11pt,a4paper]{article}
\usepackage[hyperref]{acl2017}
\usepackage{times}
\usepackage{latexsym}
\usepackage{amssymb}
\usepackage{url}
\usepackage{algorithm,algorithmic,amsmath}
\usepackage{graphicx}
\usepackage{comment}
\usepackage{mathtools}
\aclfinalcopy 

\author{
Ye Zhang$^1$ \hspace{.1\textwidth} {\bf Matthew Lease}$^2$\hspace{.1\textwidth} {\bf Byron C. Wallace}$^3$\\\\
$^1${Department of Computer Science, University of Texas at Austin}\\
$^2${School of Information, University of Texas at Austin}\\
$^3${College of Computer \& Information Science,
	Northeastern University}\\
\tt{yezhang@utexas.edu, ml@utexas.edu, byron@ccs.neu.edu}
}

\title{Exploiting Domain Knowledge via Grouped Weight Sharing \\ with Application to Text Categorization}

\date{}

\begin{document}
\maketitle
\begin{abstract}

A fundamental advantage of neural models for NLP is their ability to learn representations from scratch. However, in practice this often means ignoring existing external linguistic resources, e.g., WordNet or domain specific ontologies such as the Unified Medical Language System (UMLS). We propose a general, novel method for exploiting such resources via \emph{weight sharing}. Prior work on weight sharing in neural networks has considered it largely as a means of model compression. In contrast, we treat weight sharing as a flexible mechanism for incorporating prior knowledge into neural models. We show that this approach consistently 
yields improved performance on classification tasks compared to baseline strategies that do not exploit weight sharing.

\end{abstract}
\section{Introduction}
\label{section:intro}
\vspace{-.5em}
Neural models are powerful in part due to their ability to learn good representations of raw textual inputs, mitigating the need for extensive task-specific feature engineering \cite{Collobert:11}. However, a downside of learning from scratch is failing to capitalize on prior linguistic or semantic knowledge, often encoded in existing resources such as ontologies. Such prior knowledge can be particularly valuable when estimating highly flexible models. In this work, we address how to exploit known relationships between words when training neural models for NLP tasks. 

We propose exploiting the \emph{feature-hashing} trick, originally proposed as a means of neural network compression \cite{chen2015}. Here we instead view the partial parameter sharing induced by feature hashing as a flexible mechanism for tying together network node weights that we believe to be similar \emph{a priori}. In effect, this acts as a regularizer that constrains the model to learn weights that agree with the domain knowledge codified in external resources like ontologies. 

\begin{figure}
\centering  
\includegraphics [width=0.475\textwidth,]{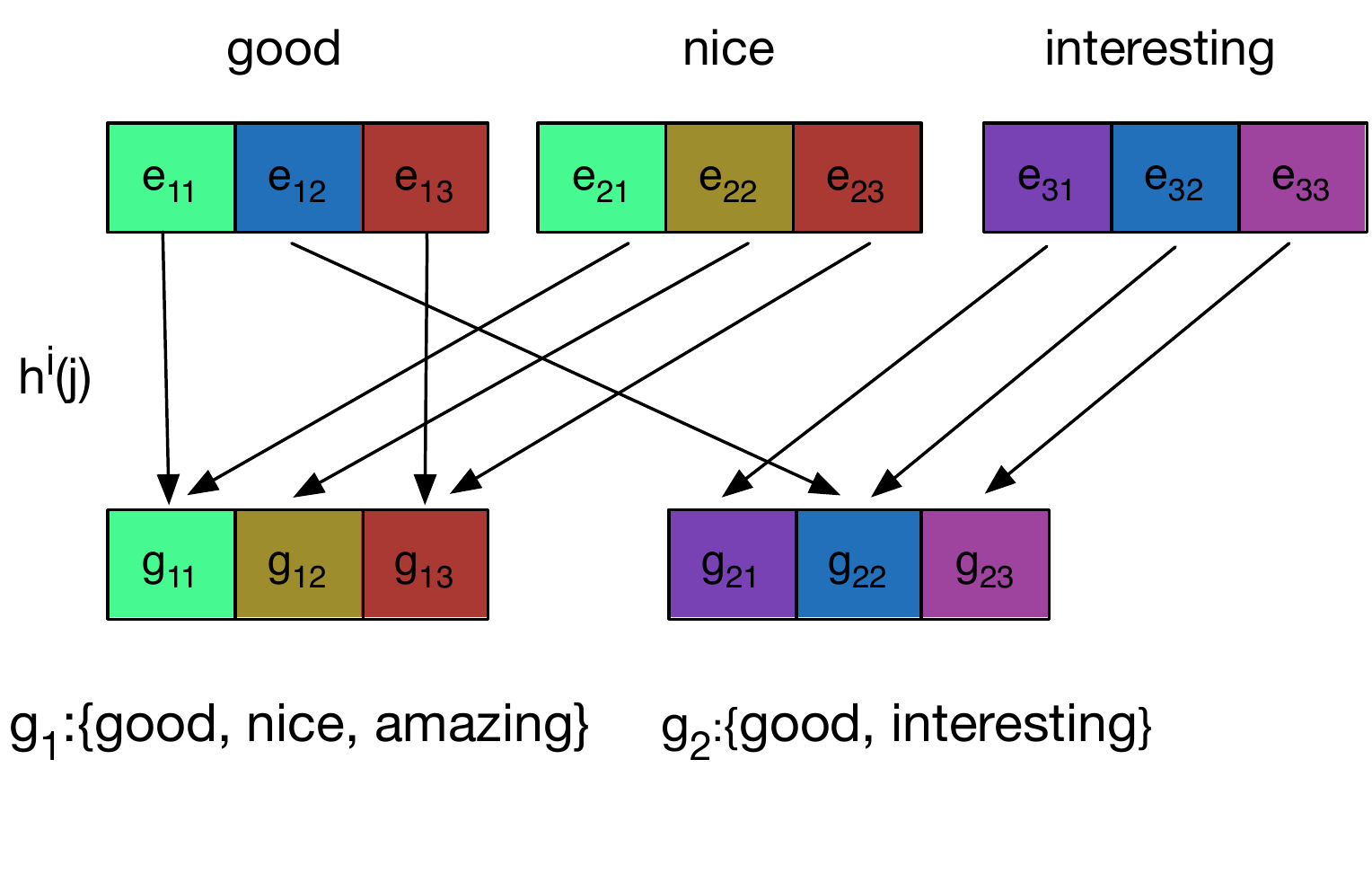}
\vspace{-2em}
\caption{An example of grouped partial weight sharing. Here there are two groups. We stochastically select embedding weights to be shared between words belonging to the same group(s).} 
 \label{fig:weight_sharing_emb}
\vspace{-1em} 
\end{figure}

More specifically, as external resources we use Brown clusters \cite{Brown:92}, WordNet \cite{miller:95} and the Unified Medical Language System (UMLS) \cite{bodenreider2004unified}. From these we derive groups of words with similar meaning. We then use feature hashing to share a subset of weights between the embeddings of words that belong to the same semantic group(s). This forces the model to respect prior domain knowledge, insofar as words similar under a given ontology are compelled to have similar embeddings.




Our contribution is a novel, simple and flexible method for injecting domain knowledge into neural models via stochastic weight sharing. Results on seven diverse classification tasks (three sentiment and four biomedical) show that our method consistently improves performance over (1) baselines that fail to capitalize on domain knowledge, and (2) an approach that uses \emph{retrofitting} \cite{faruqui2014retrofitting} as a preprocessing step to encode domain knowledge prior to training.



\section{Grouped Weight Sharing}
\label{section:methods}

We incorporate similarity relations codified in existing resources (here derived from Brown clusters, SentiWordNet and the UMLS) as prior knowledge in a Convolutional Neural Network (CNN).\footnote{The idea of sharing weights to reflect known similarity is general and could be applied with other neural architectures.} To achieve this we construct a shared embedding matrix such that words known \emph{a priori} to be similar are constrained to share some fraction of embedding weights. 

Concretely, suppose we have $N$ groups of words derived from an external resource. Note that one could derive such groups in several ways; e.g., using the synsets in SentiWordNet. We denote groups by $\{g_1,g_2, ..., g_N\}$. Each group is associated with an embedding $\mathbf{g}_{g_i}$, which we initialize by averaging the pre-trained embeddings of each word in the group.

To exploit both grouped and independent word weights, we adopt a two-channel CNN model \cite{zhang2016mgnc}. The embedding matrix of the first channel is initialized with pre-trained word vectors. We denote this input by $\mathbf{E}^p\in \mathbb{R}^{V\times d}$ ($V$ is the vocabulary size and $d$ the dimension of the word embeddings). The second channel input matrix is initialized with our proposed weight-sharing embedding $\mathbf{E}^s\in \mathbb{R}^{V\times d}$. $\mathbf{E}^s$ is initialized by drawing from both $\mathbf{E}^p$ and the external resource following the process we describe below. 

\begin{figure}
\centering  
\includegraphics [width=0.425\textwidth,]{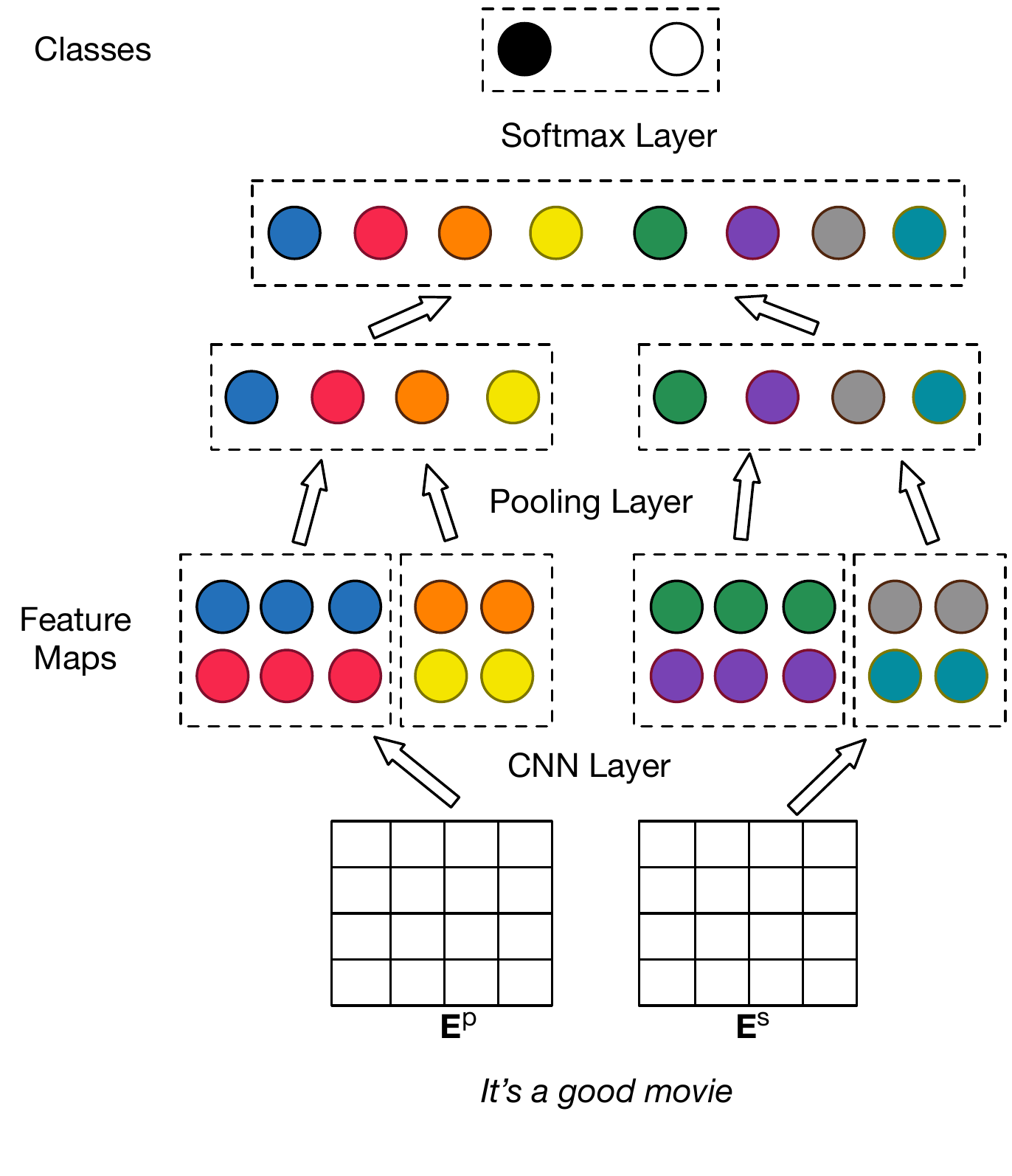}
\vspace{-1.5em}
\caption{Proposed two-channel model. The first channel input is a standard pre-trained embedding matrix. The second channel receives a partially shared embedding matrix constructed using external linguistic resources.} 
 \label{fig:two-channel CNN}
 \vspace{-1.1em}
\end{figure}

Given an input text sequence of length $l$, we construct sequence embedding representations  $\mathbf{W}^p\in \mathbb{R}^{l\times d}$ and $\mathbf{W}^s \in \mathbb{R}^{l\times d}$ using the corresponding embedding matrices. We then apply independent sets of linear convolution filters on these two matrices. Each filter will generate a feature map vector $\mathbf{v}\in\mathbb{R}^{l-h+1}$ ($h$ is the filter height).
We perform 1-max pooling over each $\mathbf{v}$, extracting one scalar per feature map. Finally, we concatenate scalars from all of the feature maps (from both channels) into a feature vector which is fed to a softmax function to predict the label (Figure \ref{fig:two-channel CNN}).

We initialize $\mathbf{E}^s$ as follows. Each row $\mathbf{e}_i\in \mathbb{R}^d$ of $\mathbf{E}_s$ is the embedding of word $i$. Words may belong to one or more groups. A mapping function $G(i)$ retrieves the groups that word $i$ belongs to, i.e., $G(i)$ returns a subset of $\{g_1,g_2, ..., g_N\}$, which we denote by $\{g^{(i)}_1,g^{(i)}_2 ... g^{(i)}_K\}$, where $K$ is the number of groups that contain word $i$.  To initialize $\mathbf{E}^s$, for each dimension $j$ of each word embedding $\mathbf{e}_i$, we use a hash function $h^i$ to map (hash) the index $j$ to one of the $K$ group IDs: 
$h^i:\mathbb{N}\rightarrow \{g^{(i)}_1,g^{(i)}_2...g^{(i)}_K\}$. 
Following \cite{weinberger2009feature,shi2009hash}, we use a second hash function $b$ 
  to remove bias induced by hashing. This is a signing function, i.e., it maps $(i,j)$ tuples to $\{+1,-1\}$
  \footnote{Empirically, we found that using this signing function does not affect performance.}.  We then set $\mathbf{e}_{i,j}$ to the product of $\mathbf{g}_{h^i(j),j}$ and $b(i, j)$. $h$ and $b$ are both approximately uniform hash functions. 
\textbf{Algorithm} \ref{alg:ini} provides the full initialization procedure.

\begin{algorithm}
 \caption{Initialization of $\mathbf{E}^s$}
  \label{alg:ini}
  \begin{algorithmic}[1]
    \FOR{$i$ in $\{1,\dots,V\}$}  
    \STATE $\{g_1^{(i)},g_2^{(i)}, \dots, g_K^{(i)}\}\coloneqq G(i)$. 
      \FOR{$j \in \{1, \dots, d\}$}
   
       \STATE 
      
      
       
       $\mathbf{e}_{i,j}\coloneqq \mathbf{g}_{h^i(j),j} \cdot b(i,j)$
      \ENDFOR
    \ENDFOR
  \end{algorithmic}
\end{algorithm}

For illustration, consider Figure \ref{fig:weight_sharing_emb}. 
Here $g_1$ contains three words: \emph{good}, \emph{nice} and \emph{amazing}, while $g_2$ has two words: \emph{good} and \emph{interesting}. The group embeddings $\mathbf{g}_{g_1}$, $\mathbf{g}_{g_2}$ are initialized as averages over the pre-trained embeddings of the words they comprise. Here, embedding parameters $\mathbf{e}_{1,1}$ and $\mathbf{e}_{2,1}$ are both mapped to $\mathbf{g}_{g_1,1}$, and thus share this value. Similarly, $\mathbf{e}_{1,3}$ and $\mathbf{e}_{2,3}$ will share value at $\mathbf{g}_{g_1,3}$. We have elided the second hash function $b$ from this figure for simplicity.


During training, we update $\mathbf{E}^p$ as usual using back-propagation \cite{RumelhartHintonWIlliams1986}. We update $\mathbf{E}^s$ and group embeddings $\mathbf{g}$ in a manner similar to \citeauthor{chen2015}\ \shortcite{chen2015}.  
In the forward propagation before each training step (mini-batch), we derive the value of $\mathbf{e}_{i,j}$ from $\mathbf{g}$: 
\begin{equation}
\label{eq:forward}
\mathbf{e}_{i,j}\coloneqq \mathbf{g}_{h^i(j),j} * b(i,j)
\end{equation}

We use this newly updated $\mathbf{e}_{i,j}$ to perform forward propagation in the CNN.

During backward propagation, we first compute the gradient of $\mathbf{E}^s$, and then we use this to derive the gradient w.r.t $\mathbf{g}s$. To do this, for each dimension $j$ in $\mathbf{g}_{g_k}$, we aggregate the gradients w.r.t $\mathbf{E}^s$ whose elements are mapped to this dimension:
\begin{equation} 
\label{eq:backward}
\nabla \mathbf{g}_{g_k,j} \coloneqq  \sum _{(i,j)}\nabla \mathbf{E}^s_{i,j} \cdot \delta_{h^i(j)=g_k} \cdot b(i,j)
\end{equation}
where $\delta_{h^i(j)=g_k}=1$ when $h^i(j)=g_k$, and 0 otherwise. Each training step involves executing Equations $\ref{eq:forward}$ and $\ref{eq:backward}$. Once the shared gradient is calculated, gradient descent proceeds as usual. We update all parameters aside from the shared weights in the standard way.

The number of parameters in our approach scales linearly with the number of channels. But the gradients can actually be back-propagated in a distributed way for each channel, since the convolutional  and embedding layers are independent across these. Thus training time scales approximately linearly with the number of parameters in one channel (if the gradient is back-propagated in a distributed way). 


\section{Experimental Setup}
\subsection{Datasets}
We use three sentiment datasets: a movie review (\textbf{MR}) dataset \cite{Pang+Lee:05a}\footnote{\url{www.cs.cornell.edu/people/pabo/movie-review-data/}}; a customer review (\textbf{CR}) dataset \cite{hu2004mining}\footnote{\url{www.cs.uic.edu/\~liub/FBS/sentiment-analysis.html}}; and an opinion dataset (\textbf{MPQA}) \cite{wiebe2005annotating}\footnote{\url{mpqa.cs.pitt.edu/corpora/mpqa\_corpus/}}.

We also use four biomedical datasets, which concern \emph{systematic reviews}. The task here is to classify published articles describing clinical trials as \emph{relevant} or \emph{not} to a well-specified clinical question. Articles deemed relevant are included in the corresponding review, which is a synthesis of all pertinent evidence \cite{wallace2010semi}. We use data from reviews that concerned:
clopidogrel (CL) for cardiovascular conditions \cite{dahabreh2013testing}; biomarkers for assessing iron deficiency in anemia (AN) experienced by patients with kidney disease \cite{chung2012biomarkers}; statins (ST) \cite{cohen2006reducing}; and proton beam (PB) therapy \cite{proton}.

\begin{table*}
\centering
\begin{tabular}{ c | c c c c } 
 & \emph{total \#instances} & \emph{vocabulary size} & \emph{\#positive instances} & \emph{\#negative instances} \\
MR & 10662 & 18765 & 5331 & 5331 \\
CR & 3773 & 5340 & 2406 & 1367 \\
MPQA & 10604 & 6246 & 3311 & 7293  \\
AN & 5653 & 5554 & 653 & 5000 \\
CL & 8288 & 3684 & 768 & 7520 \\
ST & 3464 & 2965 & 173 & 3291 \\
PB & 4749 & 3086 & 243 & 4506  \\
\end{tabular}
\caption{Corpora statistics.}
\label{statistics}
\end{table*}

\subsection{Implementation Details and Baselines}

We use SentiWordNet \cite{baccianella2010sentiwordnet}\footnote{\url{sentiwordnet.isti.cnr.it}} for the sentiment tasks. 
SentiWordNet assigns to each synset of wordnet three sentiment scores: positivity, negativity and objectivity, constrained to sum to 1. We keep only the synsets with positivity or negativity scores greater than 0, i.e., we remove synsets deemed objective. 
The synsets in SentiWordNet constitute our groups.
We also use the Brown clustering algorithm\footnote{\url{github.com/percyliang/brown-cluster}} on the three sentiment datasets. We generate 1000 clusters and treat each as a group. 

For the biomedical datasets, we use the Medical Subject Headings (MeSH) terms\footnote{\url{www.nlm.nih.gov/bsd/disted/meshtutorial/}} attached to each abstract to classify them. Each MeSH term has a tree number indicating the path from the root in the UMLS. For example, `Alagille Syndrome' has tree number `C06.552.150.125'; periods denote tree splits, numbers are nodes. We induce groups comprising MeSH terms that share the same first three parent nodes, e.g., all terms with `C06.552.150' as their tree number prefix constitute one group.

We compare our approach to several baselines. All use pre-trained embeddings to initialize $\mathbf{E}^p$, but we explore several approaches to exploiting $\mathbf{E}^s$: (1) randomly initialize $\mathbf{E}^s$; (2)
initialize $\mathbf{E}^s$ to reflect the group embedding $\mathbf{g}$, but do not share weights during the training process, i.e., do not constrain their weights to be equal when we perform back-propagation;  
(3) use linguistic resources to \emph{retro-fit} \cite{faruqui2014retrofitting} the pre-trained embeddings, and use these to initialize $\mathbf{E}^s$. 
For \emph{retro-fitting}, we first construct a graph derived from SentiWordNet. Then we run belief-propagation on the graph to encourage linked words to have similar vectors. This is a pre-processing step only; we do not impose weight sharing constraints during training.

For the sentiment datasets we use three filter heights (3,4,5) for each of the two CNN channels. For the biomedical datasets, we use only one filter height (1), because the inputs are unstructured MeSH terms.\footnote{For this work we are ignoring title and abstract texts.} In both cases we use 100 filters of each unique height. For the sentiment datasets, we use Google word2vec \cite{mikolov2013efficient}\footnote{\url{code.google.com/archive/p/word2vec/}} to initialize $\mathbf{E}^p$. For the biomedical datasets, we use word2vec trained on biomedical texts \cite{moen2013distributional}\footnote{\url{bio.nlplab.org/}} to initialize $\mathbf{E}^p$.
For parameter estimation, we use Adadelta \cite{Zeiler12adadelta:an}. 
Because the biomedical datasets are imbalanced, we use downsampling ~\cite{zhang2016rationale,zhang2015sensitivity} to effectively train on balanced subsets of the data. 

We developed our approach using the MR sentiment dataset, tuning our approach to constructing groups from the available resources -- experiments on other sentiment datasets were run after we finalized the model and hyperparameters. Similarly, we used the anemia (AN) review as a development set for the biomedical tasks, especially w.r.t.\ constructing groups from MeSH terms using UMLS.


\section{Results}
\label{section:results}
We replicate each experiment five times (each is a 10-fold cross validation), and report the mean (min, max) across these replications. Results on the sentiment and biomedical corpora in are presented in Tables \ref{table::senti} and \ref{table::bio}, respectively.\footnote{Sentiment task results are not directly comparable to prior work due to different preprocessing steps.} These exploit different external resources to induce the word groupings that in turn inform weight sharing. We report AUC for the biomedical datasets because these are highly imbalanced (see Table \ref{statistics}).

Our method
improves performance compared to all relevant baselines (including an approach that also exploits external knowledge via retrofitting) in six of seven cases. Informing weight \emph{initialization} using external resources improves performance independently, but additional gains are realized by also enforcing sharing during training.




\begin{table*}
\centering
\begin{tabular}{ c | c c c } 
 \emph{Method} & \emph{MR} & \emph{CR} & \emph{MPQA} \\
  p only & 81.02 (80.84,81.24) & 84.34 (84.21,84.53) & 89.41 (89.22,89.58)  \\ 
 p + r & 81.25 (81.19,81.32) & 84.33 (84.24,84.38) & 89.63 (89.58,89.71)\\
 p + retro & 81.35 (81.23,81.51) & 84.16 (84.09,84.28) & 89.61 (89.48,89.77) \\
 p + S (no sharing) & 81.39 (81.32,81.43) & 84.13 (84.06,84.21) & 89.71 (89.67,89.75) \\
 p + B (no sharing) & 81.50 (81.29,81.63) & 84.60 (84.53,84.66) & 89.57 (89.52,89.61) \\
 p + S (sharing) & 81.69 (81.60,81.78) & 84.34 (84.24,84.43) &  89.84 (89.74,90.13)\\
 p + B (sharing) & \textbf{81.83 (81.80,81.87)} & \textbf{84.68 (84.64,84.72)} & \textbf{89.97 (89.74,90.13)} \\
\end{tabular}
\caption{Accuracy mean (min, max) on sentiment datasets.  `p': channel initialized with the pre-trained embeddings $\mathbf{E}^p$. `r': channel randomly initialized. 
`retro': initialized with retofitted embeddings.
`S/B (no sharing)': channel initialized with $\mathbf{E}^s$ (using SentiWordNet or Brown clusters), but weights are not shared during training. 
`S/B (sharing)': proposed weight-sharing method.}
\label{table::senti}
\end{table*}



\begin{table*}
\centering
\begin{tabular}{ c | c c c c} 
\emph{Method} & \emph{AN} & \emph{CL} & \emph{ST} & \emph{PB} \\
p only & 86.63 (86.57,86.67) & 88.73 (88.51,89.00) & 67.15 (66.00, 67.91) & 90.11 (89.46, 91.03)\\
p + r & 85.67 (85.46,85.95) & 88.87 (88.56,89.03) & 67.72 (67.65,67.86) & 90.12 (89.87,90.47) \\
p + retro & 86.46 (86.32,86.65) & 89.27 (88.89,90.01) & \textbf{67.78 (67.56,68.00)} & 90.07 (89.92,90.20) \\
p + U  & 86.60 (86.32,87.01) & 88.93 (88.67,89.13) & 67.78 (67.71,67.85) & 90.23 (89.84,90.47) \\
p + U(s)  & \textbf{87.15 (87.00,87.29)} & \textbf{89.29 (89.09,89.51)} & 67.73 (67.58,67.88) &\textbf{90.99 (90.46,91.59)}  \\
\end{tabular}
\caption{AUC mean (min, max) on the biomedical datasets. Abbreviations are as in Table \ref{table::senti}, except here the external resource is the UMLS MeSH ontology (`U').`U(s)' is the proposed weight sharing method utilizing ULMS.}
\label{table::bio}
\end{table*}



We note that our aim here is not necessarily to achieve state-of-art results on any given dataset, but rather to evaluate the proposed method for incorporating external linguistic resources into neural models via weight sharing. We have therefore compared to baselines that enable us to assess this. 

\section{Related Work}
\label{section:related-work}


\noindent {\bf Neural Models for NLP}. Recently there has been enormous interest in neural models for NLP generally \cite{Collobert:11,goldberg2016primer}. Most relevant to this work, simple CNN based models (which we have built on here) have proven extremely effective for text categorization \cite{kim2014convolutional,zhang2015sensitivity}. 




\vspace{0.2em}
\noindent {\bf Exploiting Linguistic Resources}. A potential drawback to learning from scratch in end-to-end neural models is a failure to capitalize on existing knowledge sources. There have been efforts to exploit such resources specifically to induce better word vectors \cite{yu2014improving,faruqui2014retrofitting,yu2016retrofitting,xu2014rc}. But these models do not attempt to exploit external resources jointly during training for a particular downstream task (which uses word embeddings as inputs), as we do here. 

Past work on sparse linear models has shown the potential of exploiting linguistic knowledge in statistical NLP models. For example, \citeauthor{Yogatama2014Linguistic}\ \shortcite{Yogatama2014Linguistic} used external resources to inform structured, grouped regularization of log-linear text classification models, yielding improvements over standard regularization approaches. Elsewhere, 
\citeauthor{doshi2014graph}\ \shortcite{doshi2014graph} proposed a variant of LDA that exploits \emph{a priori} known tree-structured relations between tokens (e.g., derived from the UMLS) in topic modeling. 


\vspace{.2em}
\noindent {\bf Weight-sharing in NNs}. Recent work has considered stochastically sharing weights in neural models. Notably, \citeauthor{chen2015}\ \shortcite{chen2015}.  
 proposed randomly sharing weights in neural networks. 
Elsewhere, \citeauthor{han2015deep}\ \shortcite{han2015deep} proposed quantized weight sharing as an intermediate step in their deep compression model. 
In these works, the primary motivation was \emph{model compression}, whereas here we view the hashing trick as a mechanism to encode domain knowledge. 

\section{Conclusion}
We have proposed a novel method for incorporating prior semantic knowledge into neural models via stochastic weight sharing. We have showed it generally improves text classification performance vs.\  model variants which do not exploit external resources and vs.\ an approach based on retrofitting prior to training. In future work, we will investigate generalizing our approach beyond classification, and to inform weight sharing using other varieties and sources of linguistic knowledge. 

{\small 
~\\\noindent
{\bf Acknowledgements.} This work was made possible by NPRP grant
NPRP 7-1313-1-245 from the Qatar National Research Fund
(a member of Qatar Foundation). The statements made
herein are solely the responsibility of the authors.
}
\vspace{-1em}

\bibliography{acl2017}

\begin{thebibliography}{}
\expandafter\ifx\csname natexlab\endcsname\relax\def\natexlab#1{#1}\fi

\bibitem[{Baccianella et~al.(2010)Baccianella, Esuli, and
  Sebastiani}]{baccianella2010sentiwordnet}
Stefano Baccianella, Andrea Esuli, and Fabrizio Sebastiani. 2010.
\newblock Sentiwordnet 3.0: An enhanced lexical resource for sentiment analysis
  and opinion mining.
\newblock In {\em LREC\/}. volume~10, pages 2200--2204.

\bibitem[{Bodenreider(2004)}]{bodenreider2004unified}
Olivier Bodenreider. 2004.
\newblock The unified medical language system (umls): integrating biomedical
  terminology.
\newblock {\em Nucleic acids research\/} 32(suppl 1):D267--D270.

\bibitem[{Brown et~al.(1992)Brown, Desouza, Mercer, Pietra, and Lai}]{Brown:92}
Peter~F Brown, Peter~V Desouza, Robert~L Mercer, Vincent J~Della Pietra, and
  Jenifer~C Lai. 1992.
\newblock Class-based n-gram models of natural language.
\newblock {\em Computational linguistics\/} 18(4):467--479.

\bibitem[{Chen et~al.(2015)Chen, Wilson, Tyree, Weinberger, and
  Chen}]{chen2015}
Wenlin Chen, James~T Wilson, Stephen Tyree, Kilian~Q Weinberger, and Yixin
  Chen. 2015.
\newblock Compressing neural networks with the hashing trick.
\newblock In {\em ICML\/}. pages 2285--2294.

\bibitem[{Chung et~al.(2012)Chung, Moorthy, Hadar, Salvi, Iovin, and
  Lau}]{chung2012biomarkers}
Mei Chung, Denish Moorthy, Nira Hadar, Priyanka Salvi, Ramon~C Iovin, and
  Joseph Lau. 2012.
\newblock {\em Biomarkers for Assessing and Managing Iron Deficiency Anemia in
  Late-Stage Chronic Kidney Disease\/}.
\newblock AHRQ Comparative Effectiveness Reviews. Agency for Healthcare
  Research and Quality (US), Rockville (MD).

\bibitem[{Cohen et~al.(2006)Cohen, Hersh, Peterson, and
  Yen}]{cohen2006reducing}
Aaron~M Cohen, William~R Hersh, K~Peterson, and Po-Yin Yen. 2006.
\newblock Reducing workload in systematic review preparation using automated
  citation classification.
\newblock {\em Journal of the American Medical Informatics Association\/}
  13(2):206--219.

\bibitem[{Collobert et~al.(2011)Collobert, Weston, Bottou, Karlen, Kavukcuoglu,
  and Kuksa}]{Collobert:11}
Ronan Collobert, Jason Weston, L{\'e}on Bottou, Michael Karlen, Koray
  Kavukcuoglu, and Pavel Kuksa. 2011.
\newblock Natural language processing (almost) from scratch.
\newblock {\em Journal of Machine Learning Research\/} 12(Aug):2493--2537.

\bibitem[{Dahabreh et~al.(2013)Dahabreh, Moorthy, Lamont, Chen, Kent, and
  Lau}]{dahabreh2013testing}
Issa~J Dahabreh, Denish Moorthy, Jenny~L Lamont, Minghua~L Chen, David~M Kent,
  and Joseph Lau. 2013.
\newblock Testing of cyp2c19 variants and platelet reactivity for guiding
  antiplatelet treatment .

\bibitem[{Doshi-Velez et~al.(2015)Doshi-Velez, Wallace, and
  Adams}]{doshi2014graph}
Finale Doshi-Velez, Byron~C Wallace, and Ryan Adams. 2015.
\newblock Graph-sparse lda: A topic model with structured sparsity.
\newblock In {\em AAAI Conference on Artificial Intelligence\/}. pages
  2575--2581.

\bibitem[{Faruqui et~al.(2014)Faruqui, Dodge, Jauhar, Dyer, Hovy, and
  Smith}]{faruqui2014retrofitting}
Manaal Faruqui, Jesse Dodge, Sujay~K Jauhar, Chris Dyer, Eduard Hovy, and
  Noah~A Smith. 2014.
\newblock Retrofitting word vectors to semantic lexicons.
\newblock {\em arXiv preprint arXiv:1411.4166\/} .

\bibitem[{Goldberg(2016)}]{goldberg2016primer}
Yoav Goldberg. 2016.
\newblock A primer on neural network models for natural language processing.
\newblock {\em Journal of Artificial Intelligence Research\/} 57:345--420.

\bibitem[{Han et~al.(2015)Han, Mao, and Dally}]{han2015deep}
Song Han, Huizi Mao, and William~J Dally. 2015.
\newblock Deep compression: Compressing deep neural networks with pruning,
  trained quantization and huffman coding.
\newblock {\em arXiv preprint arXiv:1510.00149\/} .

\bibitem[{Hu and Liu(2004)}]{hu2004mining}
Minqing Hu and Bing Liu. 2004.
\newblock Mining and summarizing customer reviews.
\newblock In {\em Proceedings of the 10th ACM SIGKDD international conference
  on Knowledge discovery and data mining\/}. pages 168--177.

\bibitem[{Kim(2014)}]{kim2014convolutional}
Yoon Kim. 2014.
\newblock Convolutional neural networks for sentence classification.
\newblock {\em arXiv preprint arXiv:1408.5882\/} .

\bibitem[{Mikolov et~al.(2013)Mikolov, Chen, Corrado, and
  Dean}]{mikolov2013efficient}
Tomas Mikolov, Kai Chen, Greg Corrado, and Jeffrey Dean. 2013.
\newblock Efficient estimation of word representations in vector space.
\newblock {\em arXiv preprint arXiv:1301.3781\/} .

\bibitem[{Miller(1995)}]{miller:95}
George~A Miller. 1995.
\newblock Wordnet: a lexical database for english.
\newblock {\em Communications of the ACM\/} 38(11):39--41.

\bibitem[{Moen and Ananiadou(2013)}]{moen2013distributional}
SPFGH Moen and Tapio Salakoski2~Sophia Ananiadou. 2013.
\newblock Distributional semantics resources for biomedical text processing.

\bibitem[{Pang and Lee(2005)}]{Pang+Lee:05a}
Bo~Pang and Lillian Lee. 2005.
\newblock Seeing stars: Exploiting class relationships for sentiment
  categorization with respect to rating scales.
\newblock In {\em Proc.\ of the ACL\/}.

\bibitem[{Rumelhart et~al.(1986)Rumelhart, Hintont, and
  Williams}]{RumelhartHintonWIlliams1986}
D.E. Rumelhart, G.E. Hintont, and R.J. Williams. 1986.
\newblock {Learning representations by back-propagating errors}.
\newblock {\em Nature\/} 323(6088):533--536.

\bibitem[{Shi et~al.(2009)Shi, Petterson, Dror, Langford, Smola, and
  Vishwanathan}]{shi2009hash}
Qinfeng Shi, James Petterson, Gideon Dror, John Langford, Alex Smola, and SVN
  Vishwanathan. 2009.
\newblock Hash kernels for structured data.
\newblock {\em Journal of Machine Learning Research\/} 10(Nov):2615--2637.

\bibitem[{Terasawa et~al.(2009)Terasawa, Dvorak, Ip, Raman, Lau, and
  Trikalinos}]{proton}
T.~Terasawa, T.~Dvorak, S.~Ip, G.~Raman, J.~Lau, and T.~A. Trikalinos. 2009.
\newblock {Charged Particle Radiation Therapy for Cancer: A Systematic Review}.
\newblock {\em Ann. Intern. Med.\/} .

\bibitem[{Wallace et~al.(2010)Wallace, Trikalinos, Lau, Brodley, and
  Schmid}]{wallace2010semi}
Byron~C Wallace, Thomas~A Trikalinos, Joseph Lau, Carla Brodley, and
  Christopher~H Schmid. 2010.
\newblock Semi-automated screening of biomedical citations for systematic
  reviews.
\newblock {\em BMC bioinformatics\/} 11(1):55.

\bibitem[{Weinberger et~al.(2009)Weinberger, Dasgupta, Langford, Smola, and
  Attenberg}]{weinberger2009feature}
Kilian Weinberger, Anirban Dasgupta, John Langford, Alex Smola, and Josh
  Attenberg. 2009.
\newblock Feature hashing for large scale multitask learning.
\newblock In {\em Proceedings of the 26th Annual International Conference on
  Machine Learning\/}. pages 1113--1120.

\bibitem[{Wiebe et~al.(2005)Wiebe, Wilson, and Cardie}]{wiebe2005annotating}
Janyce Wiebe, Theresa Wilson, and Claire Cardie. 2005.
\newblock Annotating expressions of opinions and emotions in language.
\newblock {\em Language resources and evaluation\/} 39(2):165--210.

\bibitem[{Xu et~al.(2014)Xu, Bai, Bian, Gao, Wang, Liu, and Liu}]{xu2014rc}
Chang Xu, Yalong Bai, Jiang Bian, Bin Gao, Gang Wang, Xiaoguang Liu, and
  Tie-Yan Liu. 2014.
\newblock Rc-net: A general framework for incorporating knowledge into word
  representations.
\newblock In {\em Proceedings of the 23rd ACM International Conference on
  Conference on Information and Knowledge Management\/}. ACM, pages 1219--1228.

\bibitem[{Yogatama and Smith(2014)}]{Yogatama2014Linguistic}
Dani Yogatama and Noah~A Smith. 2014.
\newblock Linguistic structured sparsity in text categorization.
\newblock In {\em Meeting of the Association for Computational Linguistics\/}.
  pages 786--796.

\bibitem[{Yu and Dredze(2014)}]{yu2014improving}
Mo~Yu and Mark Dredze. 2014.
\newblock Improving lexical embeddings with semantic knowledge.
\newblock In {\em ACL\/}. pages 545--550.

\bibitem[{Yu et~al.(2016)Yu, Cohen, Bernstam, and Wallace}]{yu2016retrofitting}
Zhiguo Yu, Trevor Cohen, Elmer~V Bernstam, and Byron~C Wallace. 2016.
\newblock Retrofitting word vectors of mesh terms to improve semantic
  similarity measures.
\newblock {\em Intl.\ Workshop on Health Text Mining and Information Analysis
  at EMNLP\/} pages 43--51.

\bibitem[{Zeiler(2012)}]{Zeiler12adadelta:an}
Matthew~D. Zeiler. 2012.
\newblock Adadelta: An adaptive learning rate method.

\bibitem[{Zhang et~al.(2016{\natexlab{a}})Zhang, Marshall, and
  Wallace}]{zhang2016rationale}
Ye~Zhang, Iain Marshall, and Byron~C Wallace. 2016{\natexlab{a}}.
\newblock Rationale-augmented convolutional neural networks for text
  classification.
\newblock {\em arXiv preprint arXiv:1605.04469\/} .

\bibitem[{Zhang et~al.(2016{\natexlab{b}})Zhang, Roller, and
  Wallace}]{zhang2016mgnc}
Ye~Zhang, Stephen Roller, and Byron Wallace. 2016{\natexlab{b}}.
\newblock Mgnc-cnn: A simple approach to exploiting multiple word embeddings
  for sentence classification.
\newblock {\em arXiv preprint arXiv:1603.00968\/} .

\bibitem[{Zhang and Wallace(2015)}]{zhang2015sensitivity}
Ye~Zhang and Byron~C Wallace. 2015.
\newblock A sensitivity analysis of (and practitioners' guide to) convolutional
  neural networks for sentence classification.
\newblock {\em arXiv preprint arXiv:1510.03820\/} .

\end{thebibliography}
\bibliographystyle{acl_natbib}

\end{document}